\documentclass[letterpaper, 10 pt, conference]{ieeeconf}  
\usepackage[utf8]{inputenc}
\usepackage{siunitx}
\usepackage{graphicx}
\usepackage{amsmath}
\usepackage{amsfonts,amssymb}
\usepackage{hyperref}
\usepackage{booktabs}
\usepackage{multirow}
\usepackage{bm}
\usepackage{tikz}
\usetikzlibrary{bayesnet}

\DeclareMathOperator{\Dehom}{D}
\DeclareSIUnit\px{px}
\DeclareMathOperator*{\argmax}{arg\,max\>}

\DeclareMathOperator{\E}{\mathbb{E}}

\newcommand\narroweq{\mkern1.5mu{=}\mkern1.5mu}
\newcommand{\bigCI}{\mathrel{\text{\scalebox{1.07}{$\perp\mkern-10mu\perp$}}}}
 
\IEEEoverridecommandlockouts                              

\title{\LARGE \bf Enhanced Human-Machine Interaction by Combining Proximity Sensing with Global Perception}

\author{Christoph Heindl$^{*}$, Markus Ikeda$^{*}$, Gernot St\"ubl$^{*}$, Andreas Pichler$^{*}$ and Josef Scharinger$^{**}$
\thanks{$^{*}$Visual Computing and Robotics, PROFACTOR GmbH, Austria
	{\tt\small christoph.heindl@profactor.at}}%
\thanks{$^{**}$Institute of Computational Perception, Johannes Kepler University, Austria
	{\tt\small josef.scharinger@jku.at}}%
}

\begin{document}

\maketitle
\thispagestyle{empty} 

\ieeefootline{\small{\emph{2\textsuperscript{nd} Workshop on Proximity Perception}\\ 2019 IEEE/RSJ International Conference on Intelligent Robots and Systems (IROS 2019), Macau, China\\
Published under the Creative Commons license (CC BY-NC-ND) at KITopen}}


\begin{abstract}
The raise of collaborative robotics has led to wide range of sensor technologies to detect human-machine interactions: at short distances, proximity sensors detect nontactile gestures virtually occlusion-free, while at medium distances, active depth sensors are frequently used to infer human intentions. We describe an optical system for large workspaces to capture human pose based on a single panoramic color camera. Despite the two-dimensional input, our system is able to predict metric 3D pose information over larger field of views than would be possible with active depth measurement cameras. We merge posture context with proximity perception to reduce occlusions and improve accuracy at long distances. We demonstrate the capabilities of our system in two use cases involving multiple humans and robots.
\end{abstract}

\section{INTRODUCTION}

Proximity perception is an active research field, aiming to equip robotics with nontactile near-field sensors to advance robot autonomy and human-machine interoperability. While proximity sensing is a compelling concept on close-up range, it fails to recognize spatio-temporal events on moderate distances from the robot \cite{hughes2018robotic}. These events, however, provide important information to create a more robust and intuitive set of interaction patterns. 

In previous works, optical systems were increasingly used to close this information gap.
Many approaches integrate active depth cameras to detect human key points and measure distances between objects in the environment. The downsides of active depth sensors are the short operating range, the sensitivity to extraneous light and the increased occlusion caused by the camera-projector arrangement.

\begin{figure}[t]  
	\centering
	\includegraphics[width=\columnwidth]{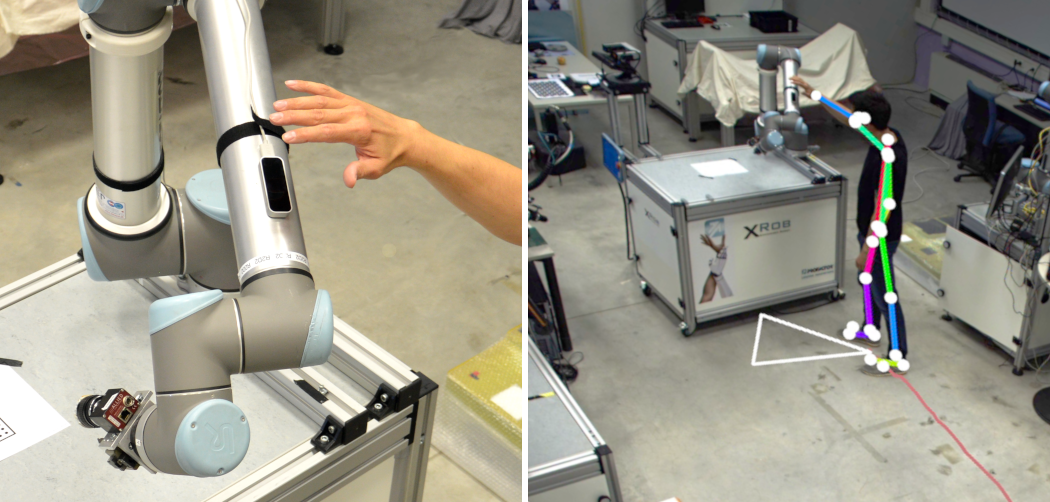}
	\caption{\label{fig:overview} Our system complements proximity sensing (left) with global perception from a bird's-eye perspective in real-time (right). Based on a color input image, we estimate the following features shown superimposed in the right image: human pose (dots connected by lines), worker orientation (white triangle) and the recent movement trajectory (red path on floor). The majority of these measurements are available in a metric 3D space and combined with proximity perception to coordinate human-machine interactions.}
\end{figure}

We describe an optical system that complements proximity perception by observing human and machines from a bird's-eye perspective (see Figure~\ref{fig:overview}). Even though our system works solely with color images from one single panoramic camera, it delivers pose information for humans and robots in a metric space. We combine the provided pose information with near-field measurements from proximity sensors to robustly detect human-machine interaction patterns. We demonstrate the benefits of merging complementary input modalities in two use cases involving multiple robots and humans \footnote{\label{fn:video}Demonstration video \href{https://youtu.be/1X0xF3m36BA}{\url{https://youtu.be/1X0xF3m36BA}}}.


\subsection{Related Work}
Vision based human-machine interaction has been studied before. Guanglong et al.~\cite{du2018online} combine RGB-D cameras and inertial measurement units to detect human gestures for robot learning. Zimmermann et al.~\cite{zimmermann20183d} use depth-based devices to record human instructions for teaching a service robot. 


Systems based on depth cameras are also used to study human-robot safety aspects. Fabrizio et al.~\cite{fabrizio2016real} describe a depth sensing enabled device to compute distances to dynamic obstacles for collision avoidance. {\v{S}}varn{\`y} et al.~\cite{petr2018} propose using 2D keypoint detection merged with RGB-D data to measure distances between a robot and a single human operator.

\subsection{Contributions}
The proposed system has a number of benefits. Our system works with a readily available single wide-angle color camera, eliminating the range limitations of active depth devices. The inference step builds on recent advances in pose estimation, enabling our system to answer complex image related queries robustly. In addition to instantaneous pose, we track individual humans to provide pose trajectories over time. Although the detection step takes place in image space, the majority of predictions are elevated to a metric 3D space, enabling natural fusion with near-field sensors to create a richer set of human-machine interactions. 

\section{METHOD}
\begin{figure}[t] 
	\centering
	\includegraphics[width=0.85\columnwidth]{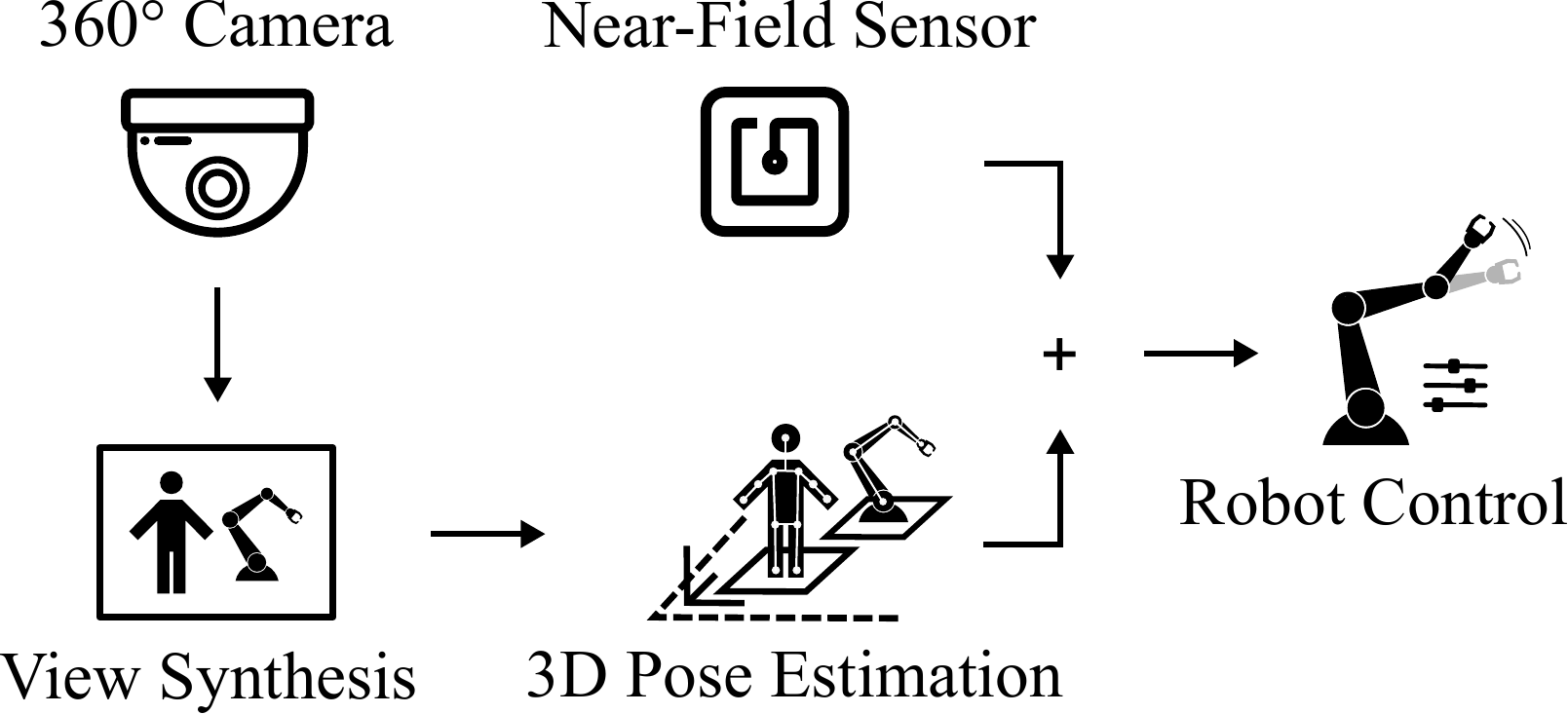}
	\caption{\label{fig:workflow} System Overview. Our approach considers panoramic images as input; these are synthesized to form one or more rectilinear views. Deeply learned neural networks then predict 2D human and robot pose keypoints. These detections are subsequently lifted to a 3D metric space using homographies of planes. Next, our system fuses near-field measurements with human/robot pose context to create location-aware events. These events in turn lead to environmental reactions defined by the application scenario.}
\end{figure}

Figure~\ref{fig:workflow} illustrates the core components of our system. These are detailed in the following sub-sections.

\subsection{View Synthesis}
\begin{figure}[htp]
	\centering
	\includegraphics[width=0.85\columnwidth]{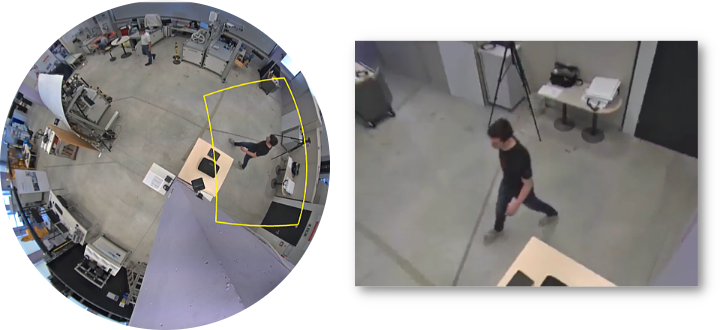}
	\caption{\label{fig:panoramic} View synthesis from panoramic color images. A virtual rectilinear camera image (right) is created from a highly distorted spherical image region (left).}
\end{figure}

In order to process panoramic color images, we first synthesize one or more synthetic rectilinear views (see Figure~\ref{fig:panoramic}). View synthesis assumes a virtual pinhole camera $P$ that is potentially rotated with respect to the physical source frame $S$. Next, every virtual camera pixel $\mathbf{u}^P$ is mapped to a corresponding source pixel $\mathbf{u}^S$ using the method described in \cite{heindl2018}. The computed pixel position is bilinearly interpolated to determine the final color value associated with $\mathbf{u}^P$.

\subsection{Pose Estimation}
\begin{figure}[h]
	\centering
	\includegraphics[width=0.85\columnwidth]{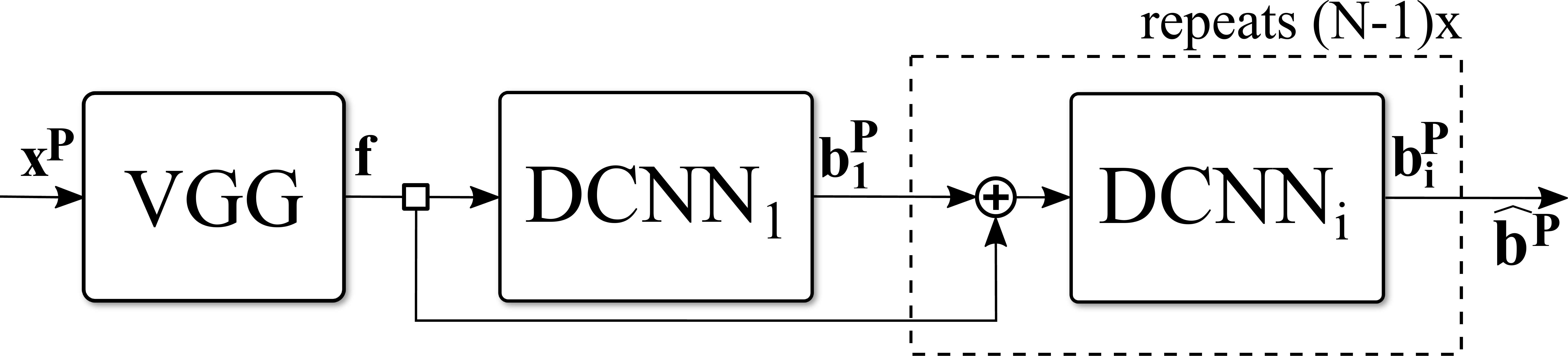}
	\caption{\label{fig:joints} Joint localization architecture \cite{cheind2019disp}. First, base features $\mathbf{f}$ are extracted from input image $\mathbf{x}^P$. An initial Deep Convolutional Neural Network (DCNN) performs initial joint belief prediction $\mathbf{b}^P_1$. A series of DCNNs then refines belief prediction to generate the final prediction $\hat{\mathbf{b}}^P$.}
  \end{figure}

To compute the 2D pose of humans and robots, we first predict 2D keypoints from synthesized color views (see Figure~\ref{fig:usecases} b,d). We use neural network architectures, depicted in Figure~\ref{fig:joints} based on works of Cao et al.~\cite{cao2017realtime} and Heindl et al.~\cite{cheind2019disp} to perform keypoint localization. Each network, composed of a series of Deep Convolutional Neural Networks (DCNNs), regresses keypoint coordinates by transforming the input color images $\mathbf{x}^P \in \mathbb{R}^{3 \times H\times W}$ into keypoint belief maps $\hat{\mathbf{b}}^P \in \mathbb{R}^{C \times H \times W}$. Here $H,W$ are image height and width and $C$ is the number of output keypoints (6 for robots, 25 for humans). Each belief map encodes the likelihood of observing a particular keypoint in a specific image region. We train these networks with both real \cite{cao2017realtime} and artificially generated images \cite{cheind2019disp}.

In a subsequent step, 2D poses are transformed into a metric space via a set of homographies. In particular, we propose the use of an image-to-ground homography $\mathbf{H}_P^G \in \mathbb{R}^{3 \times 3}$, to map image positions to metric ground coordinates as follows
\begin{equation}
  \mathbf{u}^G = \Dehom\left(\mathbf{H}_P^G\begin{bmatrix}
    u^P_x,
    u^P_y,
    1
  \end{bmatrix}^T \right) \label{eq:conv},
\end{equation}
where $\Dehom(\cdot)$ is the dehomogenization operator $\begin{bmatrix}x',y'\end{bmatrix} = \begin{bmatrix}x/z, y/z\end{bmatrix}$. Because such mappings are accurate only for body parts sufficiently close to the ground (such as foot positions), we use a statistical body model to gain the ability to map extra keypoints such as hips and shoulders. These additional points serve to predict body orientation and to stabilize body positions in case of partial occlusions. Regarding robots, we map only the base joints to determine their location.

\subsection{Multiple Object Tracking} 
We filter and track human trajectories using a Kalman filter, assuming a linear dynamic motion model (see Figure~\ref{fig:overview}, \ref{fig:usecases}d). All operations are performed in ground plane coordinates to avoid perspective effects. To assign newly detected poses to existing ones, we create a bipartite graph: Given a set of pose detections $\mathcal{D}^t$ and a set of forward predicted poses $\mathcal{P}^t$ at time $t$, we add an edge $e_{dp}$ for every possible combination of detected $d \in \mathcal{D}^t$ and tracked $p \in P^t$ poses to the graph. The cost $C_{dp}$ associated with edge $e_{dp}$ is computed as the Euclidean distance between the ground positions of $d$ and $p$. We use the method of Kuhn et al.~\cite{kuhn1955hungarian} to solve for the optimal assignments and update the Kalman filters correspondingly.

\subsection{Merging Pose Information with Proximity Measurements}
\label{sec:merge}
Pose information alone has already proven useful in our experiments with human-robot interaction. However, line-of-sight occlusions and large object distances limit the system's ability to measure small movements accurately. Therefore, we merge global pose context with gestures detected by a proximity device\footnote{\texttt{https://www.leapmotion.com/}}. 

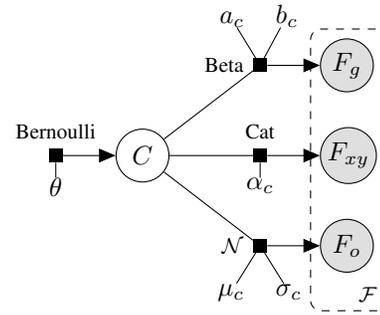
\begin{figure}[htp]
	\centering
	\begin{tikzpicture}
		\node[latent] (c) {$C$};
		\node[obs, right=of c, yshift=1.2cm, xshift=1cm] (fg) {$F_g$};
		\node[obs, right=of c, yshift=0cm, xshift=1cm] (fxy) {$F_{xy}$};
		\node[obs, right=of c, yshift=-1.2cm, xshift=1cm] (fo) {$F_o$};

		\plate[dashed] {plate1} {(fg) (fxy) (fo)} {$\mathcal{F}$}; 
		
		\factor[left=of c, xshift=-0.3cm] {fac-theta-c} {Bernoulli} {} {};    
		\node[const, below=of fac-theta-c, yshift=0.8cm]  (theta) {$\theta$}; %
		\factoredge {theta} {fac-theta-c} {c};

		\factor[left=of fg, xshift=-0.3cm] {fac-ab-fg} {left:Beta} {} {};    
		\node[const, above left=of fac-ab-fg, xshift=0.6cm, yshift=-0.3cm]  (a) {$a_c$}; %
		\node[const, above right=of fac-ab-fg, xshift=-0.6cm, yshift=-0.3cm]  (b) {$b_c$}; %
		\factoredge {c, a,b} {fac-ab-fg} {fg};

		\factor[left=of fo, xshift=-0.3cm] {fac-musig-fo} {left:$\mathcal{N}$} {} {};    
		\node[const, below left=of fac-musig-fo, xshift=0.6cm, yshift=0.3cm]  (mu) {$\mu_c$}; %
		\node[const, below right=of fac-musig-fo, xshift=-0.6cm, yshift=0.3cm]  (sig) {$\sigma_c$}; %
		\factoredge {c,mu,sig} {fac-musig-fo} {fo};
	
		\factor[left=of fxy, xshift=-0.3cm] {fac-alpha-fxy} {Cat} {} {};    
		\node[const, below=of fac-alpha-fxy, yshift=0.8cm]  (alpha) {$\alpha_c$}; %
		\factoredge {c,alpha} {fac-alpha-fxy} {fxy};
	
	\end{tikzpicture}
	
	\caption{\label{fig:fusionmodel} Probabilistic data fusion. The presence of a command intended for robot $r$ is modelled as a binary latent random variable $C$. Our model jointly considers a) gesture device confidence $F_g$, b) position of human $F_{xy}$ and c) orientation of operator $F_o$ to predict the posterior probability of $C$.}
\end{figure}

In particular, we treat data fusion as a probabilistic classification problem and consider the presence of gestures intended for robot $r$ to be latent random variables $C \in \{0,1\}$. At each time-step we observe the following noisy features: a) a confidence level of the (possibly carried) proximity sensor for an active gesture ${F_g \in [0,1]}$, b) the relative position of the operator closest to the robot in ground plane coordinates $F_{xy} \in \mathbb{R}^2$ and c) the orientation of the operator with respect to the robot, expressed as the cosine of the angle $F_o \in [-1,1]$. Given the observations $\mathcal{F}=\{F_g, F_{xy}, F_o\}$, we compute the posterior probability of the presence of a command ${\textrm{p}(C \mid \mathcal{F})}$ using Bayes theorem $$\textrm{p}(C \mid \mathcal{F}) = \frac{\textrm{p}(\mathcal{F} \mid C)\textrm{p}(C)}{\textrm{p}(\mathcal{F})}.$$ For computational reasons, we introduce the following independence assumptions $$(f \bigCI g \mid C) \quad \forall f,g \in \mathcal{F}, f \neq g.$$
That is, all observed features are independent from each other given the state of command $C$. Thus, the posterior probability simplifies to $$\textrm{p}(C \mid \mathcal{F}) = \frac{1}{Z}\; \textrm{p}(C)\prod_{f \in \mathcal{F}}\textrm{p}(f \mid C),$$ where $Z$ is the partition function given by $$Z = \sum_{c \in \{0,1\}} \textrm{p}(C \narroweq c)\prod_{f \in \mathcal{F}}\textrm{p}(f \mid C \narroweq c).$$ Our model, shown graphically in Figure~\ref{fig:fusionmodel}, follows the structure of a na\"ive Bayes classifier. We assume the following underlying probability distributions
\begin{align*}
	 C &\sim \textrm{Bernoulli}(\theta)\\
	 F_g \mid C \narroweq c &\sim \textrm{Beta}(a_c, b_c)\\
	 F_{xy} \mid C \narroweq c &\sim \textrm{Categorical}(\alpha_c)\\
	 F_o \mid C \narroweq c &\sim \textrm{Normal}(\mu_c, \sigma_c).
\end{align*}
Here, $\textrm{Categorical}(\alpha_c)$ refers to a two-dimensional distribution with bin probabilities $\alpha_c$. We have chosen to model the conditional probability $\textrm{p}(F_o \mid C)$ using a normal distribution for practical reasons. Strictly speaking, using a normal distribution is an improper assumption because its support is $(-\infty, \infty)$ and not $[0,1]$.
However, probabilities beyond $\pm 3\sigma$ quickly drop to zero and therefore do not pose a problem for our use case.

The parameters of our model are estimated in a semi-supervised fashion. That is, we consider $\mathcal{F}$ to be observed for all training samples. In addition, we observe $C$ for a fraction of training samples. The parameters of each distribution are estimated by maximizing the joint likelihood of fully observed and partially observed samples using a variant of expectation maximization (see Appendix~\ref{sec:semisupervisedem} for details). 

\section{EVALUATION}

\subsection{Experimental Setup}
We perform all experiments in an environment that covers the volume of \SI[product-units=single]{10 x 8 x 3.5}{\metre} using a single color camera with resolution \SI[product-units=single]{2464x2056}{\px}, mounted \SI{3.5}{\metre} above ground. We estimate the ground plane homography using a chessboard object. Our setup consists of two robots: an UR10 and a KUKA iiwa, both of which are placed in close proximity to each other (see Figure~\ref{fig:usecases}a). We use a standard computing unit equipped with a single NVIDIA GeForce GTX 1080 Ti for pose estimation at \SI{15}{\hertz}. The gesture sensing device is connected to a portable Laptop. ZeroMQ\footnote{\url{https://zeromq.org/}} is used to exchange messages between all computing entities.

\subsection{Pose Estimation Accuracy}
In this work we consider metric accuracies, as image accuracies have been reported previously \cite{heindl2018, cheind2019disp}. Assuming an error-free intrinsic camera matrix, the uncertainty in detecting point correspondences for homography estimation is measured to be $\sigma_C=\SI{3.5}{\px}$. The parameter uncertainties of $\mathbf{H}_P^G$ are estimated using the method of Criminsi et al.~\cite{criminisi1999plane}. The uncertainty of 2D keypoint detection is $\sigma_P=\SI{15}{\px}$. We transform input uncertainties to measurement uncertainties by propagating errors through Equation~\ref{eq:conv} to the first order. Table~\ref{tab:uncertainties} lists measurement uncertainties as a function of object-to-camera distance.
\begin{table}[h]
	\centering
	\begin{tabular}{lr}
		\toprule
		Distance &  Uncertainty \\
		\midrule
		\SI{3}{\metre}  & \SI{0.05}{\metre} \\
		\SI{10}{\metre}  & \SI{0.30}{\metre} \\
		\bottomrule
	\end{tabular}
	\caption{\label{tab:uncertainties} Measurement uncertainties as a function of object distance.}
	\vspace{-6mm}
\end{table}

\subsection{Classification Accuracy}
\begin{figure}[htp]
	\centering
	\includegraphics[width=0.95\columnwidth]{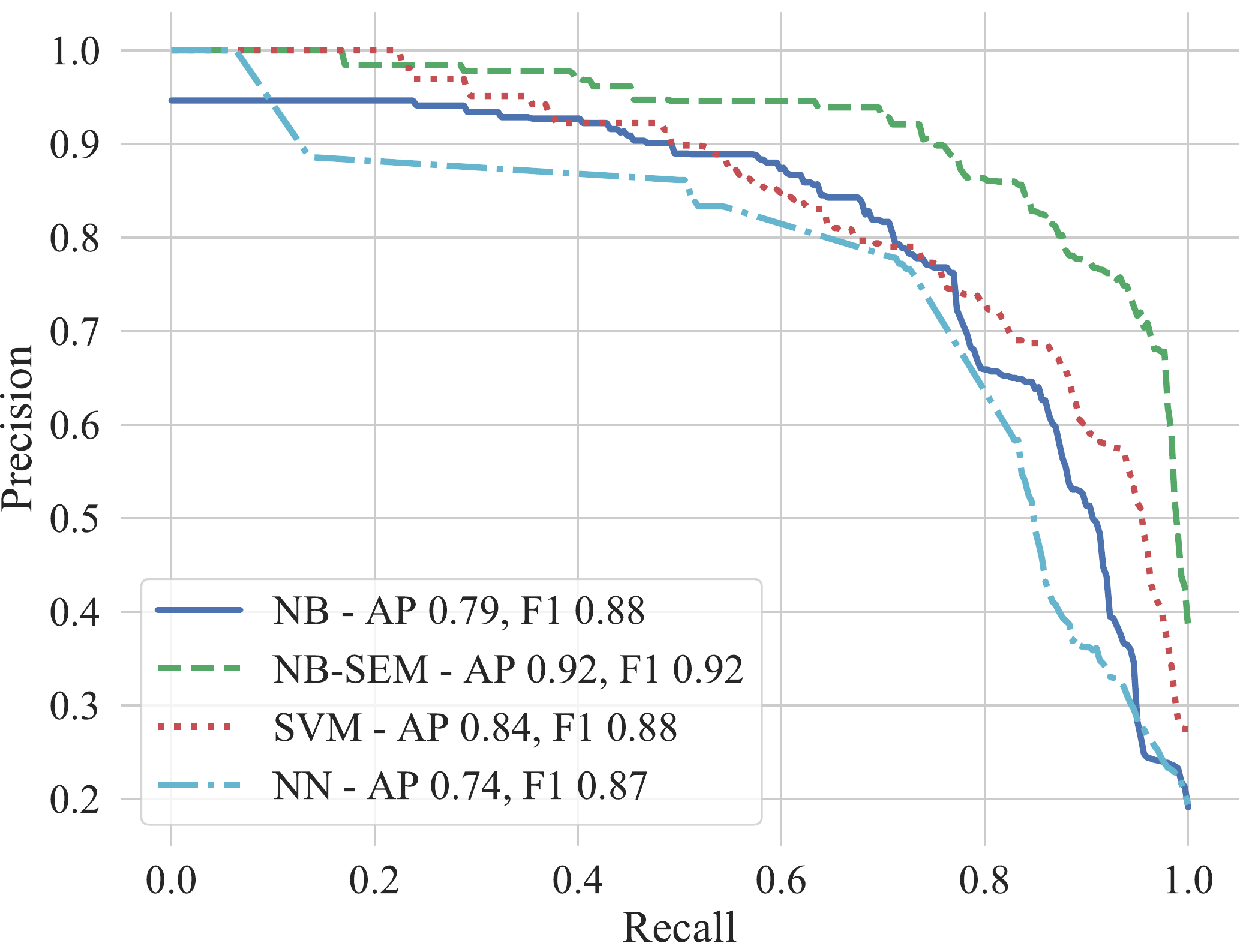}
	\caption{\label{fig:evalfusion} Gesture classification performance. Each curve represents the precision/recall of one of the following classifiers: na\"ive Bayes (NB), na\"ive Bayes trained with unlabelled data (NB-SEM), support vector machine (SVM), shallow neural network (NN). All classifiers are trained on \SI{2}{\percent} of annotated data. Only NB-SEM additionally makes use of unlabeled data. Also shown: Average Precision (AP) and macro F1 score.}
	\vspace{-0.4cm}
\end{figure}

To evaluate the data fusion approach, we conduct several experiments using the robot orchestration use case (see Section~\ref{sec:demo}). For training we record 4500 observations of $\mathcal{F}$. This corresponds to \SI{5}{\minute} of possible human-robot interaction. In a downstream step, we manually label $C$ based on visual inspection. We then train several types of classifiers assuming a fully observed (FO), i.e. $\mathcal{F} \cup \{C\}$, to partially observed (PO), i.e. just $\mathcal{F}$, ratio of \SI{2}{\percent}. Figure~\ref{fig:evalfusion} shows the advantages of our proposed na\"ive Bayes method using semi-supervised training compared to a) na\"ive Bayes without unlabeled data usage, b) SVM and c) a neural network (Appendix~\ref{sec:classifierparams} lists hyper-parameters). All classifiers are evaluated on a separate fully annotated test set of 5000 samples. 

\section{DEMONSTRATIONS}
\label{sec:demo}
\begin{figure}[htp]
	\centering
	\includegraphics[width=0.95\columnwidth]{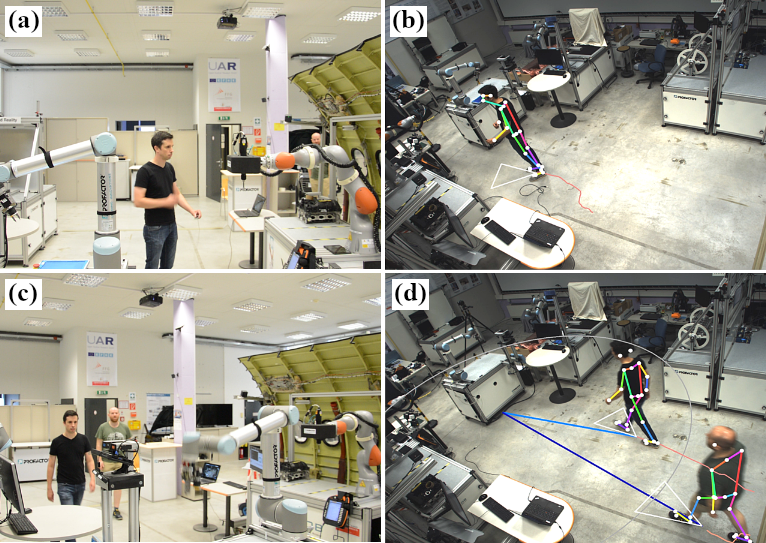}
	\caption{\label{fig:usecases} Use Cases (UCs); (a,b) UC1 Robot orchestration - operator's position and orientation selects the intended robot to receive near-field gesture commands; (c,d) UC2 Robot Interaction - adjusting robot speed in human presence for near-field interaction.}
\end{figure}

We demonstrate the interaction potential of our system in two use cases (UCs)\footnote{Video link is provided in footnote~\ref{fn:video}.}:
\begin{description}
  \item[UC1] We orchestrate multiple robots by placing a near-field gesture-sensitive device in the hands of an operator. The operator's position and orientation with respect to the robots is used to predict the intended receiver of gesture commands (see Figure~\ref{fig:usecases} a,b).
  \item[UC2] We show an adaptive robot speed control to simplify human-machine cooperation. As humans approach, we automatically decelerate the robot. Likewise, we accelerate the robot to full operating speed as humans leave (see Figure~\ref{fig:usecases} c,d). 
\end{description}

\section{CONCLUSION AND DISCUSSION}
We present a single monocular camera-based system to provide real-time pose detection for human-machine interaction at large scales. Albeit our system does not sense depth directly, we develop a homography based solution to lift pose predictions to a metric 3D space. Furthermore, we show that merging global vision context and proximity data in a probabilistic framework helps to maintain robust detection over long distances. Finally, we present the results of a semi-supervised learning approach to data fusion that increases classification accuracy when only a few marked training samples are available. 




\bibliographystyle{unsrt}
\bibliography{aeye}

\appendix
\subsection{Semi-Supervised Expectation Maximization}
\label{sec:semisupervisedem}
\begin{figure}[h]
    \centering
    \begin{tikzpicture}
        \node[obs] (z_i) {$C^i$};
        \node[obs, below=of z_i] (x_i) {$\mathcal{F}^i$};
        \node[const, right=of z_i] (theta) {$\Omega$};
        \node[latent, right=of theta] (z_j) {$\tilde{C}^j$};
		\node[obs, below=of z_j] (x_j) {$\tilde{\mathcal{F}}^j$};		

        \edge {z_i} {x_i};
        \edge {z_j} {x_j};
        \edge {theta} {z_i, x_i, z_j, x_j};
        \plate {} {(x_i)(z_i)} {$i \le N$}; 
        \plate {} {(x_j)(z_j)} {$j \le M$}; 
    \end{tikzpicture}
    \caption{\label{fig:lvm} Generative latent variable model with random variables grouped into fully and partially observed samples. Global model parameters are collected in $\Omega$.}
\end{figure}
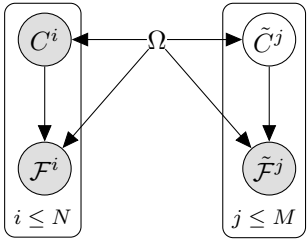

In semi-supervised training we assume that for a fraction of training samples we know the value of the latent variable $C$, while for a usually much larger fraction of data we don't.
Following the notation introduced in Section~\ref{sec:merge}, we assume to be given $N$ complete observations corresponding to the set $\{(C^i,\mathcal{F}^i)\}_{i\le N}$ of random variables. In addition $M$ partial observations of $\{\tilde{\mathcal{F}}^j\}_{j\le M}$ are provided. All observations are considered to be independent and identically distributed. The model's global parameters are collected in $\Omega=\{\theta, a_c, b_c, \alpha_c, \mu_c, \sigma_c\}$. Figure~\ref{fig:lvm} shows a graphical representation illustrating the situation.

The joint probability given $\Omega$ factors as follows
\begin{equation}
\begin{split}
p(\bm{C},& \bm{\mathcal{F}},\bm{\tilde{C}},\bm{\mathcal{\tilde{F}}} \mid \Omega) = \\ 
& \quad \prod_{i=1}^N p(C^i,\mathcal{F}^i \mid \Omega)\prod_{j=1}^M p(\tilde{C}^j, \mathcal{\tilde{F}}^j \mid \Omega).
\end{split}
\end{equation}
where we denote $\{C^i\}_{i\le N}$ by $\bm{C}$ and do similar for the other variables. We seek to optimize $\Omega$ by maximizing the log-likelihood of the observables
$$ \Omega^* = \argmax_\Omega \log p(\bm{C}, \bm{\mathcal{F}}, \bm{\mathcal{\tilde{F}}}\mid \Omega),$$ which requires integrating over latent $\bm{\tilde{C}}$. Following the objective of semi-supervised expectation maximization~(EM)~\cite{heindl2019em} gives
\begin{align}
	(\hat q, \hat\Omega) &= \argmax_{q, \Omega}\mathcal{L}(q, \Omega) \nonumber \\
	&= \argmax_{q, \Omega} \Bigl[ \log p(\bm{C}, \bm{\mathcal{F}} \mid \Omega) \nonumber + \\
	& \qquad \E_{\bm{\tilde{C}} \sim q(\bm{\tilde{C}})} \log \frac{p(\bm{\tilde{C}}, \bm{\tilde{\mathcal{F}}} \mid \Omega)}{q(\bm{\tilde{C}})} \Bigr] \label{eq:em},
\end{align}
where $q$ is a tractable distribution over $\bm{\tilde{C}}$. Expectation maximization then iteratively optimizes for $q$ (E-step) and $\Omega$ (M-step) in an alternating scheme, until a (local) maximum is reached. At iteration $t+1$, the solution to the E-step for our model is given by
\begin{align}
	q^{t+1}(\bm{\tilde{C}}) &= p(\bm{\tilde{C}} \mid \bm{C},\bm{\mathcal{F}},\bm{\mathcal{\tilde{F}}},\Omega^t) \nonumber \\
	&= \prod_{j=1}^M p(\tilde{C}^j \mid \mathcal{\tilde{F}}^j, \Omega^t) = \prod_{j=1}^M q_j^{t+1}(\tilde{C}^j)
\end{align}
where we made use of the independence assumptions underlying our model. The E-Step is thus equal to the original EM algorithm~\cite{dempster1977maximum}. The M-step updates the parameters $\Omega$ of our model as follows 
\begin{align}
	\Omega^{t+1} &= \argmax_{\Omega}\mathcal{L}(q^{t+1}, \Omega) \nonumber \\
	&= \argmax_\Omega \Biggl[ \left(\sum_{i=1}^N \log p(C^i \vert \Omega) + \log p(\mathcal{F}^i \vert C^i, \Omega)\right) +  \nonumber \\
	& \! \sum_{j=1}^{M}\sum_{c\in\{0,1\}} q^{t+1}_j(\tilde{C}^j \narroweq c)\log p(\tilde{C}^j \narroweq c, \tilde{\mathcal{F}}^j \mid \Omega) \Biggr] \label{eq:mstep},
\end{align}
which follows from Equation~\ref{eq:em} by applying the models independence assumptions and considering $q^{t+1}(\bm{\tilde{C}})$ to be constant. In Equation~\ref{eq:mstep} the fully observed samples serve as a guidance bias for the optimization procedure. Finally, we solve for $\Omega^{t+1}$ by setting the gradient to zero $$\nabla_\Omega\, \mathcal{L}(q^{t+1}, \Omega) = \mathbf{0}.$$ At \href{https://github.com/cheind/proximity-fusion}{\footnotesize{\url{https://github.com/cheind/proximity-fusion}}} we provide source code for reproducibility.

\subsection{Additional Classification Experiments}

Table~\ref{tab:addinfo} compares the macro F1 scores of different classification approaches at varying fractions of full observations. Especially when only few annotated data points are available, our method outperforms the other classifier variants.
\begin{table}[h]
	\centering
	\begin{tabular}{lrrrr}
		\toprule
		{} &   NB &  NB-SEM &  SVM &   NN \\
		Fraction FO    &      &         &      &      \\
		\midrule
		0.02 & 0.79 &  \textbf{0.92} & 0.84 & 0.74 \\
		0.20 & 0.91 &  \textbf{0.93} & 0.91 & \textbf{0.93} \\
		0.80 & 0.92 &  \textbf{0.94} & 0.93 & \textbf{0.94} \\
		\bottomrule
	\end{tabular}
	\caption{\label{tab:addinfo} F1 macro scores of gesture classifiers at varying levels of fully observed data (FO). See Figure~\ref{fig:evalfusion} for abbreviations.}
\end{table}

\subsection{Classifier Hyper-Parameterization}
\label{sec:classifierparams}
The variants of na\"ive Bayes (NB, NB-SEM) discretize $F_{xy}$ into 4 by 4 bins. NB-SEM balances the observed/latent terms in Equation~\ref{eq:mstep}, so that the influence is independent of the number of samples. NB uses closed-form maximum likelihood estimators. Additionally, for FO samples a gain of factor 10 is used when $C^i \narroweq 1$ to account for imbalanced class distributions. NB-SEM is trained using BFGS with analytic gradients. SVM uses a RBF kernel with $\gamma \narroweq 3$ and a penalty term of 1. The NN uses 5 hidden units, ReLU activation, L2 weight regularization $\alpha \narroweq 1e-5$ and is trained using BFGS.

\end{document}